\documentclass[10pt,journal]{IEEEtran}
\usepackage{cite}
\usepackage{amsmath,amssymb}
\usepackage{graphicx}
\usepackage{booktabs}
\usepackage{tabularx}
\usepackage{array}
\usepackage[table]{xcolor}
\usepackage[hidelinks]{hyperref}
\graphicspath{{figure/}{manuscript/figure/}}
\newcolumntype{L}[1]{>{\raggedright\arraybackslash}p{#1}}
\newcolumntype{Y}{>{\raggedright\arraybackslash}X}
\definecolor{wcmblue}{RGB}{214,224,237}
\definecolor{wcmgray}{RGB}{232,233,235}
\definecolor{wcmrule}{RGB}{120,125,132}
\newcolumntype{B}[1]{>{\columncolor{wcmblue}\raggedright\arraybackslash}p{#1}}
\newcolumntype{G}[1]{>{\columncolor{wcmgray}\raggedright\arraybackslash}p{#1}}
\newcolumntype{Z}{>{\columncolor{wcmblue}\raggedright\arraybackslash}X}
\newcolumntype{Q}{>{\columncolor{wcmgray}\raggedright\arraybackslash}X}
\newcommand{\wcmtablestyle}{%
  \renewcommand{\arraystretch}{1.12}%
  \setlength{\tabcolsep}{4pt}%
  \setlength{\aboverulesep}{0pt}%
  \setlength{\belowrulesep}{0pt}%
  \setlength{\extrarowheight}{2pt}%
  \arrayrulecolor{wcmrule}%
}

\title{Neuro-Symbolic Agentic AI for Networked Low-Altitude UAVs}

\author{Yuqi Ping, Tianhao Liang, Nanchi Su, Guangyu Lei, Junwei Wu, Qinyu Zhang, and Tingting Zhang
\thanks{Yuqi Ping, Tianhao Liang, Nanchi Su, Guangyu Lei, Junwei Wu, Qinyu Zhang, and Tingting Zhang are with the College of Informatics, Harbin Institute of Technology, Shenzhen 518000, China (e-mail: pingyq@stu.hit.edu.cn, liangth@hit.edu.cn, sunanchi@hit.edu.cn, GuangyuLei@stu.hit.edu.cn, 220210419@stu.hit.edu.cn, zqy@hit.edu.cn, zhangtt@hit.edu.cn). Corresponding authors are Tingting Zhang and Tianhao Liang.}}

\begin{document}
\bstctlcite{wcm_bib_control}
\maketitle

\begin{abstract}
Networked low-altitude unmanned aerial vehicles (UAVs) need reliable and adaptive decision-making capabilities to operate under uncertain observations, dynamic environments, and intermittent connectivity, while many existing agentic systems remain limited by hallucination risks, data dependence, and weak generalization. This article investigates neuro-symbolic agentic AI (NSAAI) as a framework for combining neural grounding, symbolic reasoning, and closed-loop agentic interaction to support more reliable and adaptive UAV autonomy. We first examine its capability foundations in data efficiency, compositional generalization, continual learning, and zero-shot transfer, and then develop a reference architecture integrating task and goal management, neuro-symbolic planning, verification and metacognition, skill execution and network interaction, and shared knowledge and memory. An urban fire-inspection case implemented in LAESim illustrates how a UAV can coordinate sensing and cloud access under intermittent connectivity, reuse a verified image-delivery skill, and satisfy explicit evidence conditions before completing the mission. The results illustrate the potential of NSAAI to support reusable skills, evidence-grounded decision-making, and adaptive mission execution in networked UAV systems. We further discuss key research directions in uncertainty-aware reasoning, knowledge and skill expansion, adaptive self-monitoring, and standardized evaluation.
\end{abstract}

\begin{IEEEkeywords}
Agentic AI, neuro-symbolic AI, unmanned aerial vehicles (UAVs), autonomous decision-making, networked UAV systems.
\end{IEEEkeywords}

\section{Introduction}

With the rapid growth of the low-altitude economy, networked low-altitude unmanned aerial vehicles (UAVs) are being increasingly deployed in inspection, search and rescue, logistics, emergency response, and other applications~\cite{ping2026navigation}. As these applications expand, UAV missions are evolving from fixed-route and single-purpose operations toward complex tasks that require continuous situational awareness, online decision-making, and adaptation to dynamic environments and task requirements. This evolution requires UAV systems to transition from manual control, predefined routes, and task-specific AI modules to sustained autonomy driven by mission objectives. Recent advances in foundation models and agent technologies provide a promising basis for this transition, enabling networked UAVs to operate increasingly as autonomous agents~\cite{ping2026mllmuavswarm}.

However, enabling networked low-altitude UAVs to operate as autonomous agents presents two key challenges. First, UAVs operate in complex low-altitude environments where wireless coverage is often uneven or intermittent~\cite{ping2026handover}. Meanwhile, many agentic systems rely on large language models (LLMs) for reasoning and planning, whose hallucinations or incorrect decisions may lead to mission failure or serious safety risks. Critical decisions therefore need to be grounded in traceable evidence and verified against action preconditions and task-completion conditions. Second, the neural models underlying many current agentic systems often require substantial task-specific data and remain limited in generalizing beyond their training conditions. Such data are costly to collect through real-world UAV operations, making it impractical to cover the diverse conditions encountered in low-altitude missions.

To address these challenges, we present a neuro-symbolic agentic AI (NSAAI) architecture for networked low-altitude UAVs. NSAAI combines neural learning, symbolic reasoning, and agentic planning and execution~\cite{hakim2026neurosymbolic}. Neural components convert multimodal observations into task-relevant information. Symbolic reasoning makes task knowledge and constraints explicit, while the agentic loop uses them to guide and revise decisions as execution feedback arrives. Together, these mechanisms support verifiable decision-making and more flexible reuse of knowledge and skills under changing task conditions. The main contributions of this article are summarized as follows.
\begin{itemize}

  \item We introduce NSAAI into networked low-altitude UAVs and examine its potential from the perspectives of data efficiency, compositional generalization, continual learning, and zero-shot transfer.

  \item We present an NSAAI reference architecture and illustrate its operation through an urban fire-inspection use case, demonstrating closed-loop mission execution under coupled perception and connectivity constraints.

  \item We discuss key research directions for reliable UAV autonomy, including uncertainty-aware reasoning, knowledge and skill expansion, adaptive self-monitoring, and standardized evaluation.

\end{itemize}

\section{NSAAI for Networked Low-Altitude UAVs: Capability Foundations}

\begin{figure}[!t]
\centering
\includegraphics[width=\columnwidth]{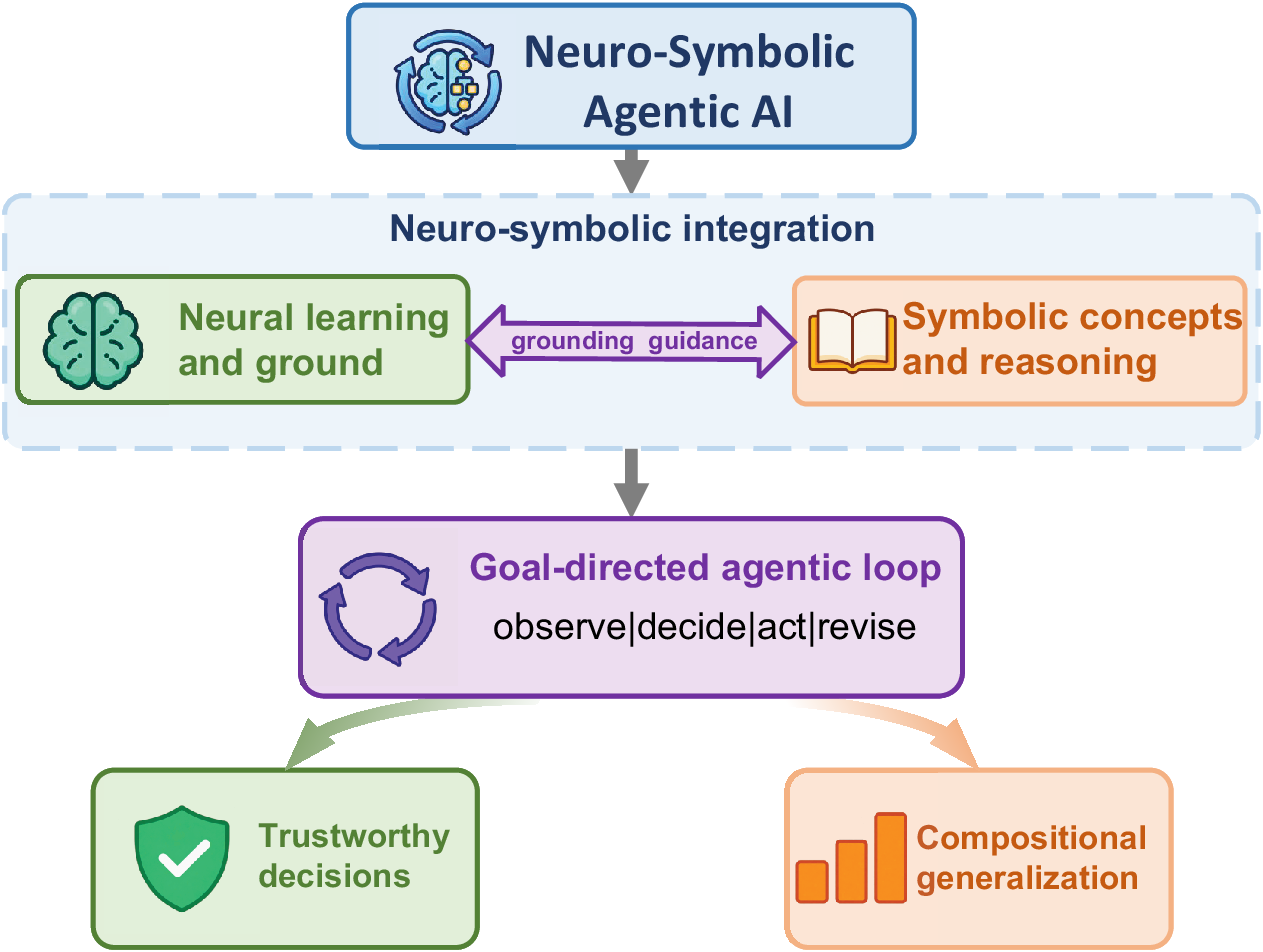}
\caption{Complementary integration of neural learning, symbolic reasoning, and agentic interaction in NSAAI.}
\label{fig:capabilities}
\end{figure}

NSAAI integrates neural learning, symbolic reasoning, and agentic planning and execution into a unified framework for autonomous UAV decision-making. Neural methods learn task-relevant representations from multimodal observations and execution experience, providing the grounding needed to interpret complex environments. However, end-to-end neural approaches often require substantial task-specific data and may struggle with systematic reasoning and generalization beyond their training conditions~\cite{mao2019concept}. Symbolic methods, in contrast, represent knowledge and action constraints explicitly, enabling structured reasoning and knowledge reuse. Their effectiveness, however, often depends on substantial domain engineering, and symbolic models are difficult to apply directly to raw observations or unknown environmental dynamics~\cite{mao2022pdsketch}. Neuro-symbolic integration brings these complementary strengths together by grounding symbolic concepts in learned representations while using symbolic structures and constraints to guide reasoning and action. Fig.~\ref{fig:capabilities} illustrates how this cooperation operates within an agentic loop driven by observation and execution feedback.

This integration has the potential to improve data efficiency, compositional generalization, continual learning, and zero-shot transfer. These capabilities describe how knowledge and skills can be learned efficiently, recombined for new situations, extended over time, and transferred across tasks. Within the agentic loop, they support adaptive decision-making as new observations and execution feedback become available. Their effectiveness depends on reliable concept grounding and the applicability of existing knowledge and skills to the current task.

\subsection{Data Efficiency}

Data efficiency concerns how much annotated data and interaction experience an agent needs to acquire useful knowledge and skills. Neuro-symbolic methods can reduce this demand by separating the learning of reusable concepts from their composition into complete tasks. Neural models learn how objects, relations, and actions correspond to observations and execution, while symbolic structures specify their dependencies and reuse across tasks. This modularity allows different components to learn from different data sources without requiring a complete mission demonstration for every combination~\cite{mao2019concept,mao2026building}.

For networked low-altitude UAVs, visual concepts may be learned from imagery, connectivity conditions from link measurements, and action skills from demonstrations or tool-interaction records. These resources can then support multiple missions through shared symbolic descriptions. The agent can also direct sensing toward a missing task condition, making new observations more relevant to the current decision. This approach is attractive where flight trials and annotations are costly, although constructing reliable concepts and skill descriptions still requires data and domain knowledge.

\begin{table*}[!t]

\caption{Capability foundations of NSAAI for networked low-altitude UAVs}

\label{tab:capability-map}

\centering

\footnotesize

\wcmtablestyle

\begin{tabularx}{\textwidth}
{B{0.19\textwidth}G{0.27\textwidth}B{0.25\textwidth}Q}

\toprule

\textbf{Capability} &
\textbf{Representative UAV need} &
\textbf{Neural role} &
\textbf{Symbolic role} \\

\midrule

Data efficiency &
Acquire reusable knowledge and skills from limited flight data and interaction experience &
Ground task-relevant concepts and skills from observations and experience &
Decompose mission knowledge into reusable concepts and skill structures \\

Compositional generalization &
Handle new missions that require unfamiliar combinations of previously learned capabilities &
Recognize familiar concepts and execute learned skills in new contexts &
Recombine existing concepts and skills according to current task requirements \\

Continual learning &
Incorporate new knowledge from mission experience while preserving previously learned capabilities &
Update concept grounding and skills from new observations and execution outcomes &
Integrate new knowledge with existing relations and constraints for future reuse \\

Zero-shot transfer &
Reuse previously learned capabilities across new missions without task-specific retraining &
Apply learned grounding and skills when they remain relevant to the new task &
Reuse existing knowledge and skill structures under new task objectives \\

\bottomrule

\end{tabularx}

\end{table*}

\subsection{Compositional Generalization}

Compositional generalization is the ability to solve tasks containing previously unseen combinations of familiar concepts, relations, and skills~\cite{chen2020ness}. Neural grounding identifies the relevant elements in observations, while symbolic composition specifies how these elements interact to satisfy a goal. Because objects, relations, and action conditions are represented separately, a new combination can be addressed through reasoning over existing knowledge.

For networked low-altitude UAVs, a new mission may require a combination of familiar concepts and skills that was not encountered during training. For example, a UAV may already know how to inspect a target, avoid obstacles, and maintain communication, but a new task may require these capabilities to be used together in a different way. By representing these concepts and skills separately, NSAAI can recombine them according to the current task instead of learning the entire mission from scratch.

\subsection{Continual Learning}

Continual learning enables an agent to incorporate new knowledge from experience while preserving previously acquired capabilities. In NSAAI, modular concept representations make such updates more localized. Neural components can learn new concept groundings from observations, while symbolic structures relate the new knowledge to what is already known and constrain how it is used. This provides a basis for incrementally extending the agent's knowledge without retraining the entire system~\cite{mei2022falcon}.

For networked low-altitude UAVs, long-term operation may reveal previously unseen targets, environmental conditions, or recurring execution patterns that are not covered by the existing knowledge base. Experience gathered during missions can be used to update the corresponding concepts or skills and make them available for future tasks. In this way, the system can progressively expand its capabilities while preserving useful knowledge acquired from earlier missions.

\subsection{Zero-Shot Transfer}

Zero-shot transfer enables previously learned knowledge or skills to be applied to a new task without additional training for the transferred capability. In NSAAI, separating reusable concept grounding and skills from task-specific objectives facilitates such transfer. Knowledge acquired for one task can therefore remain useful when a new objective requires the same concepts or capabilities in a different context~\cite{hsu2023ns3d}. Unlike compositional generalization, which focuses on new combinations of familiar elements, zero-shot transfer emphasizes the reuse of learned capabilities across tasks.

For networked low-altitude UAVs, knowledge learned in one mission can support another when the underlying concepts and skills remain applicable. For example, spatial concepts learned during inspection may later support viewpoint selection in a search task, while a previously learned communication skill may be reused whenever its operating conditions are satisfied. Such transfer allows common capabilities to be reused across missions without task-specific retraining.

Table~\ref{tab:capability-map} summarizes these four capabilities and the complementary roles of neural grounding and symbolic reasoning in supporting them. Together, they show how NSAAI can reduce data dependence, generalize through recombination, accumulate knowledge from experience, and reuse learned capabilities across missions. The next section translates these capability-level advantages into a reference architecture for networked low-altitude UAVs.

\begin{figure*}[!t]
\centering
\includegraphics[width=0.85\textwidth]{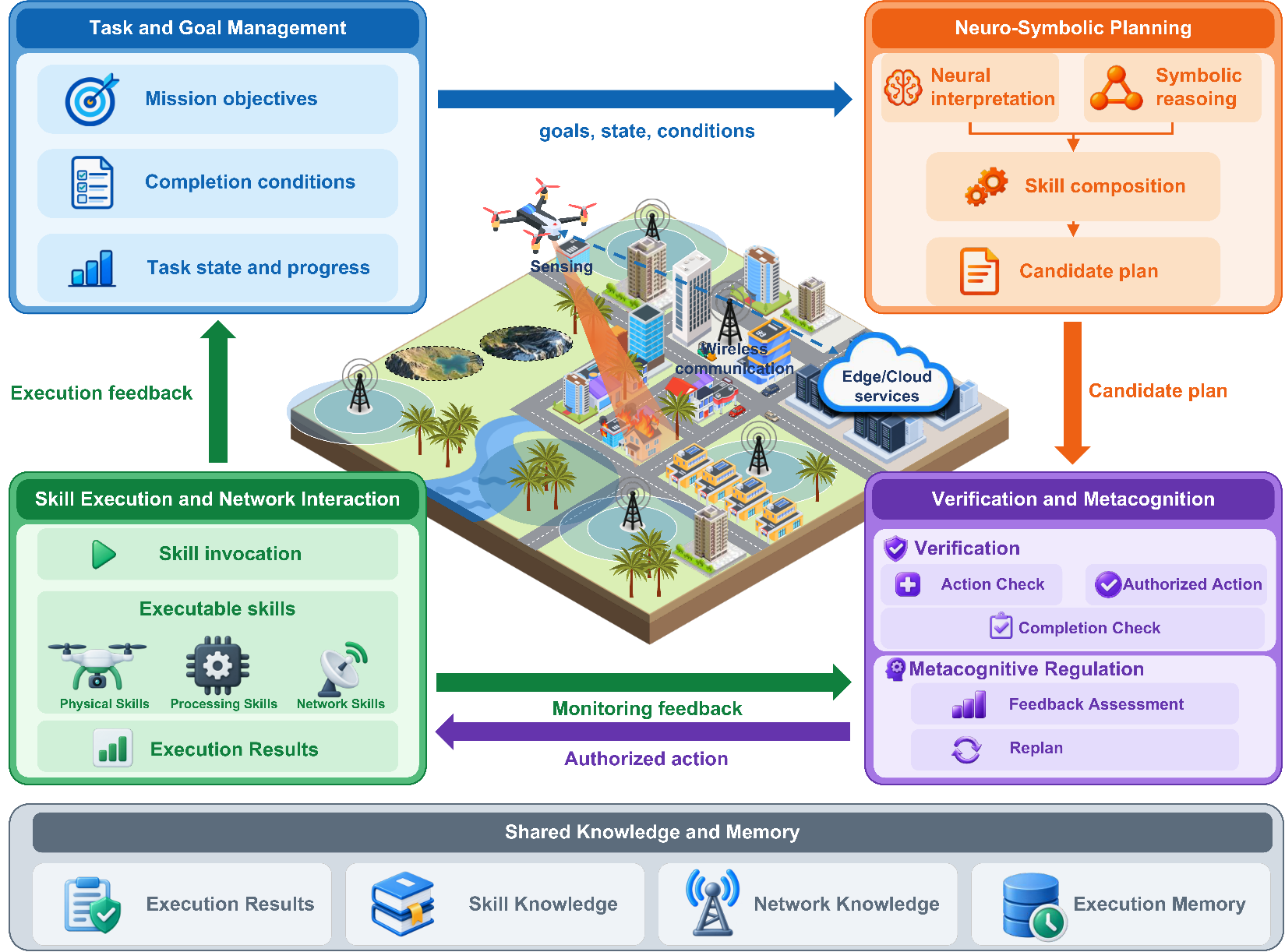}
\caption{NSAAI reference architecture for networked low-altitude UAVs.}
\label{fig:architecture}
\end{figure*}

\section{NSAAI Architecture for Networked Low-Altitude UAVs}

To incorporate the capability foundations into networked low-altitude UAV systems, we develop a reference architecture organized around four functional modules covering task and goal management, neuro-symbolic planning, verification and metacognition, and skill execution and network interaction. A shared knowledge and memory layer provides common support across these modules, as shown in Fig.~\ref{fig:architecture}.

The task and goal management module maintains mission objectives, task state, and completion conditions. The neuro-symbolic planning module generates candidate plans by combining neural interpretation with symbolic rules and skill descriptions. Before execution, the verification function checks whether candidate actions satisfy their preconditions, mission constraints, and evidence requirements. Authorized actions are then carried out by the skill execution and network interaction module, which returns observations, execution outcomes, and communication feedback. Metacognitive regulation uses this feedback to determine whether additional sensing, replanning, or strategy adaptation is required. The shared knowledge and memory layer stores rules, skills, and execution experience that support the entire decision loop.

\subsection{Task and Goal Management}

The task and goal management module maintains the mission objectives, UAV task state, completion conditions, evidence requirements, and execution progress. It integrates onboard perception, navigation state, wireless-link measurements, and responses from edge or remote services into a unified task state. The state records detected objects, UAV positions, connectivity conditions, acquired evidence, and completed or pending mission requirements, while distinguishing uncertain observations from confirmed facts and expected effects from actual outcomes.

The maintained state provides the current basis for planning and verification. Missing, uncertain, or outdated information is explicitly represented so that planning or metacognitive regulation can determine whether additional sensing, link measurement, or evidence acquisition is required. Execution and communication feedback continuously update this state so that later decisions remain consistent with the current physical and network conditions.

\subsection{Neuro-Symbolic Planning}

The neuro-symbolic planning module converts the current UAV task state and mission objectives into candidate skill sequences. Neural modules interpret visual observations, mission instructions, and other contextual inputs to identify relevant objects, states, and skills~\cite{ichter2023saycan}, while symbolic reasoning organizes these candidates according to UAV operating rules, object relations, skill preconditions, expected effects, and communication requirements.

Navigation, sensing, onboard processing, transmission, and remote-service invocation are treated as reusable skills that can be recombined as environmental and network conditions change. For example, onboard sensing and processing can continue during a link outage, while communication-dependent operations are postponed, redirected, or replaced by locally executable alternatives. The resulting skill sequence remains a candidate plan until its applicability and execution conditions are verified.

\subsection{Verification and Metacognition}

The verification and metacognition module provides supervisory evaluation and regulation over planning and execution. Verification checks candidate actions against the current task state, skill preconditions, mission and safety constraints, and available evidence. It determines whether a proposed action is authorized for execution and whether a mission or subgoal has actually satisfied its completion conditions. Following the external-verification principle of LLM-Modulo frameworks~\cite{kambhampati2024modulo}, these judgments are separated from candidate generation.

Metacognitive regulation extends this supervisory function by assessing feedback accumulated during execution and subsequent state updates. It monitors repeated link failures, inconsistent perception results, failed service calls, and mismatches between expected and observed outcomes to determine whether the current knowledge, assumptions, plan, or strategy remains appropriate. Such feedback can trigger additional sensing, replanning, or strategy adaptation rather than repeatedly executing an ineffective operation. Any resulting plan or knowledge change remains subject to the same task and safety constraints before it affects later UAV decisions.

\subsection{Skill Execution and Network Interaction}

The skill execution and network interaction module invokes actions that have passed verification, including navigation, sensing, link probing, onboard processing, data transmission, and edge or remote-service access. Flight controllers, sensing interfaces, and communication stacks execute these operations under their own real-time constraints and return structured results such as UAV position, sensor observations, link quality, acknowledgments, service responses, and execution errors.

These results are associated with the corresponding skill and task object so that the agent can determine whether the expected effect has occurred. Changes in connectivity, service availability, acknowledgment status, or environmental observations are reflected in the task state through execution feedback. The same feedback is also available to metacognitive regulation for detecting persistent failure patterns and determining whether the current plan or strategy should be revised.


\begin{figure*}[!t]

\centering

\includegraphics[width=\textwidth]{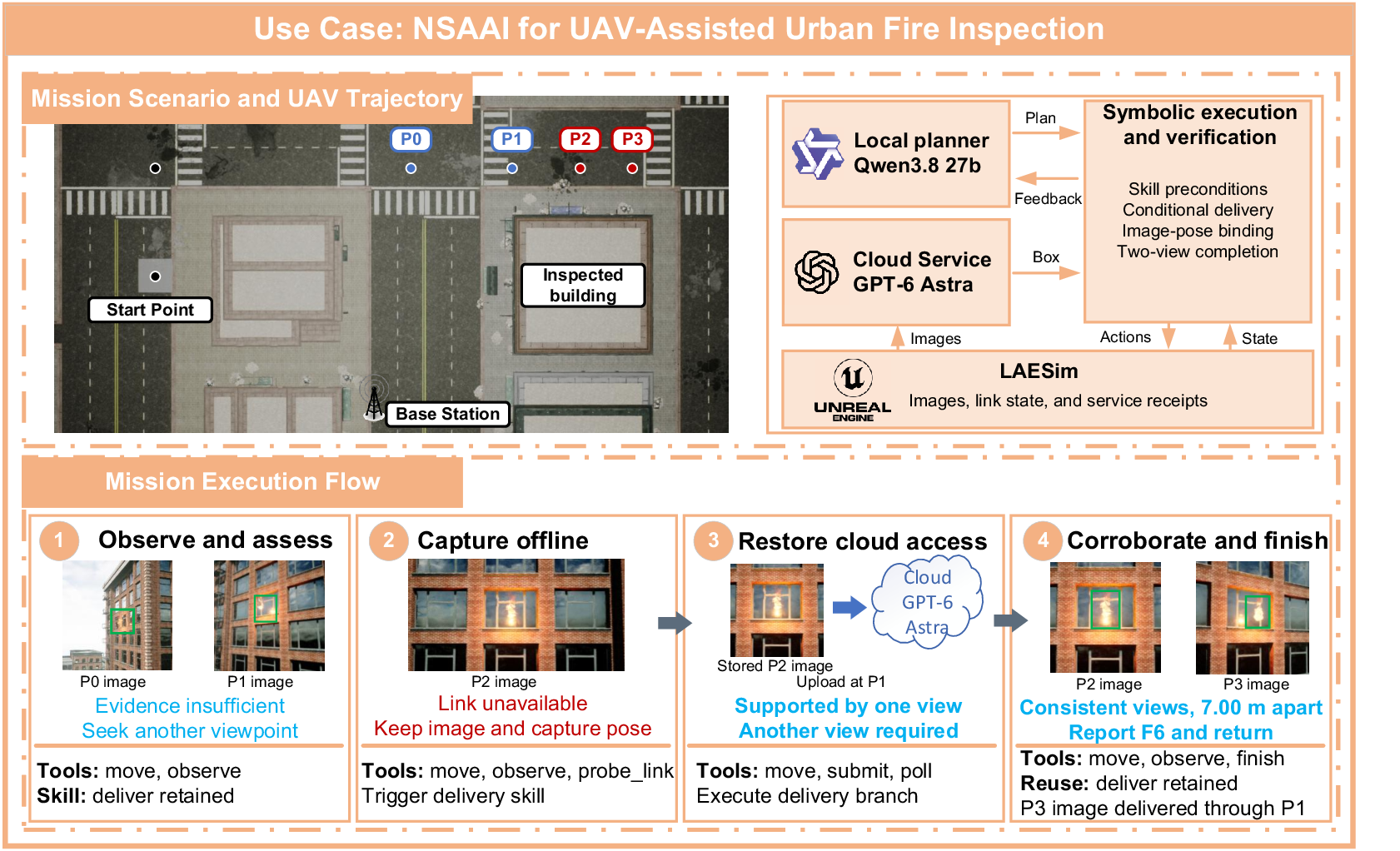}

\caption{NSAAI-assisted urban fire inspection: mission scenario, neuro-symbolic cooperation, and four stages of one online mission.}

\label{fig:usecase}

\end{figure*}

\subsection{Shared Knowledge and Memory}

The shared knowledge and memory layer stores UAV mission rules, airspace and safety constraints, skill descriptions, communication requirements, and historical execution records. It provides common references for task-state interpretation, planning, execution, and verification. While the task state represents the current mission situation, persistent execution memory records historical observations, attempted skills, connectivity conditions, acknowledgments, successes, failures, and supporting evidence accumulated over previous decision steps.

Both successful and failed executions can contribute to later adaptation. Neural modules can use accumulated feedback to propose revisions to operational rules or skill conditions. Proposed revisions are checked against recorded outcomes and protected task and safety requirements before being incorporated into the shared knowledge used for later UAV decisions. Safety constraints and mission-completion requirements are therefore not directly modified by online adaptation.

\section{Use Case: NSAAI for UAV-Assisted Urban Fire Inspection}

To illustrate the proposed NSAAI architecture in a networked low-altitude UAV mission, we consider an urban fire-inspection scenario in which perception and connectivity constraints are coupled. A UAV departs from an occluded base to inspect a reported building fire, identify the affected exterior level, and return. Useful viewpoints may lack cloud connectivity, while connected positions may provide insufficient visual evidence. The mission therefore requires the UAV to coordinate observation, communication, remote analysis, and evidence verification as conditions evolve. We first describe how the NSAAI architecture is instantiated in the simulation and then present the resulting closed-loop mission execution.

\subsection{Simulation Setup and NSAAI Instantiation}

The use case is implemented in our LAESim platform~\cite{laesim}, which provides an urban building environment with fire and smoke. Candidate viewpoints, flight corridors, and public facade geometry are available to the UAV, whereas the fire location and connectivity at unvisited positions are initially unknown. A local Qwen3.8 model with 27 billion parameters performs onboard task-level planning, while a remote GPT-6 Astra service analyzes fresh $1280\times960$ color images. Wireless connectivity is modeled using line-of-sight and communication-range constraints to determine remote-service availability.

The UAV is tasked with inspecting a reported building fire, identifying the affected exterior level, and returning to base with supporting evidence. During the mission, it must obtain observations from multiple viewpoints and use remote analysis to confirm the fire location. Useful viewpoints may not always provide connectivity, so the agent must coordinate sensing, movement, and cloud access while ensuring that the final report is supported by consistent visual evidence.

The reference architecture is instantiated through complementary neural and symbolic roles within an agentic loop. Qwen interprets the current task state and selects viewpoints or reusable skills, while the cloud-based GPT-6 Astra model extracts fire-related visual evidence from UAV images. Symbolic representations encode task conditions, connectivity states, skill requirements, and completion criteria, supporting structured execution and evidence verification. A reusable image-delivery skill is induced from short tool-interaction examples, verified, and fixed before flight~\cite{shao2026nsi}. Execution outcomes and unmet evidence conditions are fed back to the planner, which updates subsequent decisions until the task completion conditions are satisfied.

\subsection{Experimental Results}

Fig.~\ref{fig:usecase} shows one online episode rather than a prescribed mission sequence. The aerial view overlays recorded UAV positions. The UAV leaves S through street corner C and approaches the building along the inspection corridor. Cloud access is available at P0 and P1 but unavailable at the closer viewpoints P2 and P3. Images acquired at P0 and P1 are successfully analyzed, but neither identifies a unique exterior level. The planner therefore selects P2. Its failed link probe triggers the delivery skill to retain the P2 image, return to P1, and upload it without recapturing. The analysis supports F6, but one valid viewpoint is insufficient to finish.

The UAV next acquires a second observation at P3 and reuses the delivery skill through P1. The green boxes in the panels are Astra detections on the recorded images. Exterior-level labels are derived by the geometric checker, not generated by the model. The P2 and P3 observations, captured 7.00~m apart, agree on F6 and satisfy the consistency checks. The UAV reports F6 and returns to base. A separate post-run comparison with the hidden scene state confirms the reported level.

The experiment illustrates three aspects of the proposed architecture. First, insufficient evidence remains distinct from mission completion because completion is governed by explicit evidence conditions. Second, observations can be preserved across connectivity interruptions and analyzed after relocation, separating where information is acquired from where remote services are accessed. Third, the same conditional delivery skill can be reused for different images and locations without prescribing the entire mission sequence.

\section{Challenges and Research Directions}

Although NSAAI provides a promising framework for autonomous decision-making, its deployment in networked low-altitude UAVs still faces several open challenges. Future systems need to reason reliably from uncertain information, expand their knowledge as missions evolve, regulate their own decision processes, and be evaluated under consistent and realistic conditions. We highlight four research directions toward more reliable and scalable NSAAI-enabled UAV systems.

\subsection{Reliable Reasoning Under Uncertainty}

Networked low-altitude UAVs rely on visual observations and positioning information~\cite{ping2026uwb} while operating under dynamically changing network connectivity, and remote-service outputs may also be uncertain or incomplete. Current neuro-symbolic systems still lack systematic mechanisms for representing and reasoning with such uncertainty. Future NSAAI systems should therefore incorporate probabilistic inference into the grounding and reasoning process, allowing uncertain observations to be represented with confidence rather than immediately converted into deterministic symbolic facts. Probabilistic reasoning can then help the UAV compare alternative hypotheses and determine when additional sensing, communication, or verification is needed before making a decision.

\subsection{Expanding Knowledge and Skills}

Many current neuro-symbolic methods rely on predefined symbolic vocabularies or concept sets, which can be too restrictive for open and evolving environments. This limitation is particularly relevant to networked low-altitude UAVs, which may encounter previously unseen objects, environmental conditions, and mission situations that are not covered by the original knowledge representation. Promising directions include grammar-based lexicon learning, concept induction from past experience, and recent LLM-based methods that can introduce new symbolic concepts from language. For UAV systems, these approaches could support the gradual expansion of task knowledge and reusable skills as new operating conditions are encountered.

\subsection{Self-Monitoring and Adaptive Decision-Making}

NSAAI-enabled UAVs should monitor not only the external environment but also the quality of their own reasoning and the effectiveness of their current strategies. Future metacognitive mechanisms should use execution and verification feedback to assess reasoning quality, adapt strategies as task and network conditions change, allocate resources between onboard and remote processing, and determine when stored knowledge needs to be updated. Such self-assessment should also rely on external evidence, including sensor feedback and independent verification, to avoid repeatedly reinforcing incorrect internal assumptions.

\subsection{Standardized Evaluation and Benchmarking}

Progress in NSAAI for networked low-altitude UAVs requires standardized evaluation frameworks that assess not only mission performance but also the quality of neuro-symbolic integration. Future benchmarks should include tasks that jointly require perception and symbolic reasoning, together with metrics for reasoning transparency, metacognitive capability, and generalization to previously unseen missions. For low-altitude UAV systems, these evaluations should further incorporate varying physical environments and wireless conditions, while reporting agentic capabilities such as adaptability and autonomy as well as decision latency, communication overhead, and computational cost. Shared task definitions, tool interfaces, and reproducible simulation or hardware-in-the-loop settings are also needed for fair comparison across different NSAAI approaches.

\section{Conclusion}

This article presented NSAAI as a promising framework for reliable and adaptive autonomy in networked low-altitude UAVs by integrating neural grounding, symbolic reasoning, and closed-loop agentic interaction. We examined its capability foundations in data efficiency, compositional generalization, continual learning, and zero-shot transfer, and developed a reference architecture that connects task management, planning, verification, execution, and shared knowledge and memory. The urban fire-inspection case illustrated how a UAV can coordinate sensing and cloud access under intermittent connectivity, reuse verified skills, and rely on explicit evidence conditions for task completion. Future research should further address uncertainty-aware reasoning, continual knowledge and skill expansion, adaptive self-monitoring, and standardized evaluation under realistic physical and communication conditions.

\bibliographystyle{IEEEtran}
\IfFileExists{reference.bib}{%
  \bibliography{reference}%
}{%
  \bibliography{manuscript/reference}%
}
\end{document}